\documentclass[journal]{IEEEtran}
\usepackage{times}
\usepackage{amssymb}
\usepackage{helvet}
\usepackage{mathrsfs}
\usepackage{url}
\usepackage{courier}
\usepackage{booktabs,caption}
\usepackage{threeparttable}
\usepackage{graphicx}
\usepackage{booktabs}
\usepackage{url}	

\usepackage{overpic}
\usepackage{color}
\usepackage[english]{babel}
\usepackage[T1]{fontenc}
\usepackage[utf8]{inputenc}
\usepackage{multirow}
\usepackage{diagbox}
\usepackage{authblk}
\usepackage[numbers, compress]{natbib}

\usepackage{graphicx}
\usepackage{caption}
\usepackage{colortbl}

\usepackage{xspace}
\usepackage{subfig}
\usepackage{amsmath}
\usepackage{algorithm}
\usepackage{algpseudocode}
\usepackage{color}
\usepackage{wrapfig}
\usepackage{lipsum}
\usepackage{multirow}
\usepackage{hyperref}

\makeatletter
\DeclareRobustCommand\onedot{\futurelet\@let@token\@onedot}
\def\@onedot{\ifx\@let@token.\else.\null\fi\xspace}
 
\def\ie{\emph{i.e}\onedot}

\makeatother

\begin{document}



\title{Exploring Feature Representation Learning for Semi-supervised Medical Image Segmentation}

\author{Huimin Wu, Xiaomeng Li, \IEEEmembership{Member, IEEE}, and Kwang-Ting Cheng, \IEEEmembership{Fellow, IEEE}

\thanks{
H. Wu is with the Department of Computer Science and Engineering, Hong Kong University of Science and Technology, Hong Kong, SAR, China (e-mail: hwubl@connect.ust.hk). 

{X. Li is the corresponding author of this work. X. Li is with the Department of Electronic and Computer Engineering, The Hong Kong University of Science and Technology, Hong Kong, SAR, China, and also with The Hong Kong University of Science and Technology Shenzhen Research Institute, Shenzhen 518057, China (e-mail: eexmli@ust.hk).}

{K.-T. Cheng is with the Department of Electronic and Computer Engineering and Department of Computer Science and Engineering, Hong Kong University of Science and Technology, Hong Kong, SAR, China (e-mail: timcheng@ust.hk).  }

{ 
This research was partially supported by HKSAR RGC General Research Fund (GRF) \#16203319.
This work is supported in part by Foshan HKUST Projects under FSUST21-HKUST10E \& FSUST21-HKUST11E and in part by the Shenzhen Municipal Central Government Guides Local Science and Technology Development Special Funded Projects under Grant 2021Szvup139.}
}
}

%

\markboth{}%
{Shell \MakeLowercase{\textit{et al.}}: Bare Demo of IEEEtran.cls for IEEE Journals}
%


\newcommand{\para}[1]{\vspace{.05in}\noindent\textbf{#1}}
\newcommand{\reviseagain}[1]{{\color{black}{#1}}}
\newcommand{\revise}[1]{{\color{black}{#1}}}
\newcommand{\xmli}[1]{{\color{blue}{[XM: #1]}}}
\newcommand{\whm}[1]{{\color{red}{[Huimin: #1]}}}



\maketitle
\IEEEpeerreviewmaketitle
\begin{abstract}
\reviseagain{This paper presents a simple yet effective two-stage framework for semi-supervised medical image segmentation. 
Unlike prior state-of-the-art semi-supervised segmentation methods that predominantly rely on pseudo supervision directly on predictions, such as consistency regularization and pseudo labeling, 
our key insight is to explore the feature representation learning with \emph{labeled and unlabeled (i.e., pseudo labeled) images} to regularize a more compact and better-separated feature space,
which paves the way for low-density decision boundary learning and therefore
enhances the segmentation performance. 
A stage-adaptive contrastive learning method is proposed, containing a boundary-aware contrastive loss that takes advantage of the \emph{labeled images} in the first stage,
as well as a prototype-aware contrastive loss to optimize both \emph{labeled and pseudo labeled images} in the second stage.
To obtain more accurate prototype estimation, which plays a critical role in prototype-aware contrastive learning,
we present an aleatoric uncertainty-aware method, namely AUA, to generate higher-quality pseudo labels.
AUA adaptively regularizes prediction consistency by taking advantage of image ambiguity, which, given its significance, is under-explored by existing works.
Our method achieves the best results on three public medical image segmentation benchmarks. 
}
\end{abstract}


\section{Introduction}
Medical image segmentation is a foundational task for computer-aided diagnosis and computer-aided surgery. In recent years, considerable efforts have been devoted to designing neural networks for medical image segmentation,  such as U-Net~\cite{ronneberger2015u}, DenseUNet~\cite{li2018h}, nnUNet~\cite{isensee2020nnu}, HyperDenseNet~\cite{dolz2018hyperdense}. However, training these models requires a large number of labeled images.
Unlike natural images, the professional expertise required for pixel-wise manual annotation of medical images makes such labeling tasks challenging and time-consuming, resulting in the difficulty of obtaining a large labeled dataset. 
Hence, semi-supervised learning, which enables training using labeled and unlabeled data, becomes an active research area for medical image segmentation. 

\begin{figure}[t]
\centering
\includegraphics[width=0.48\textwidth]{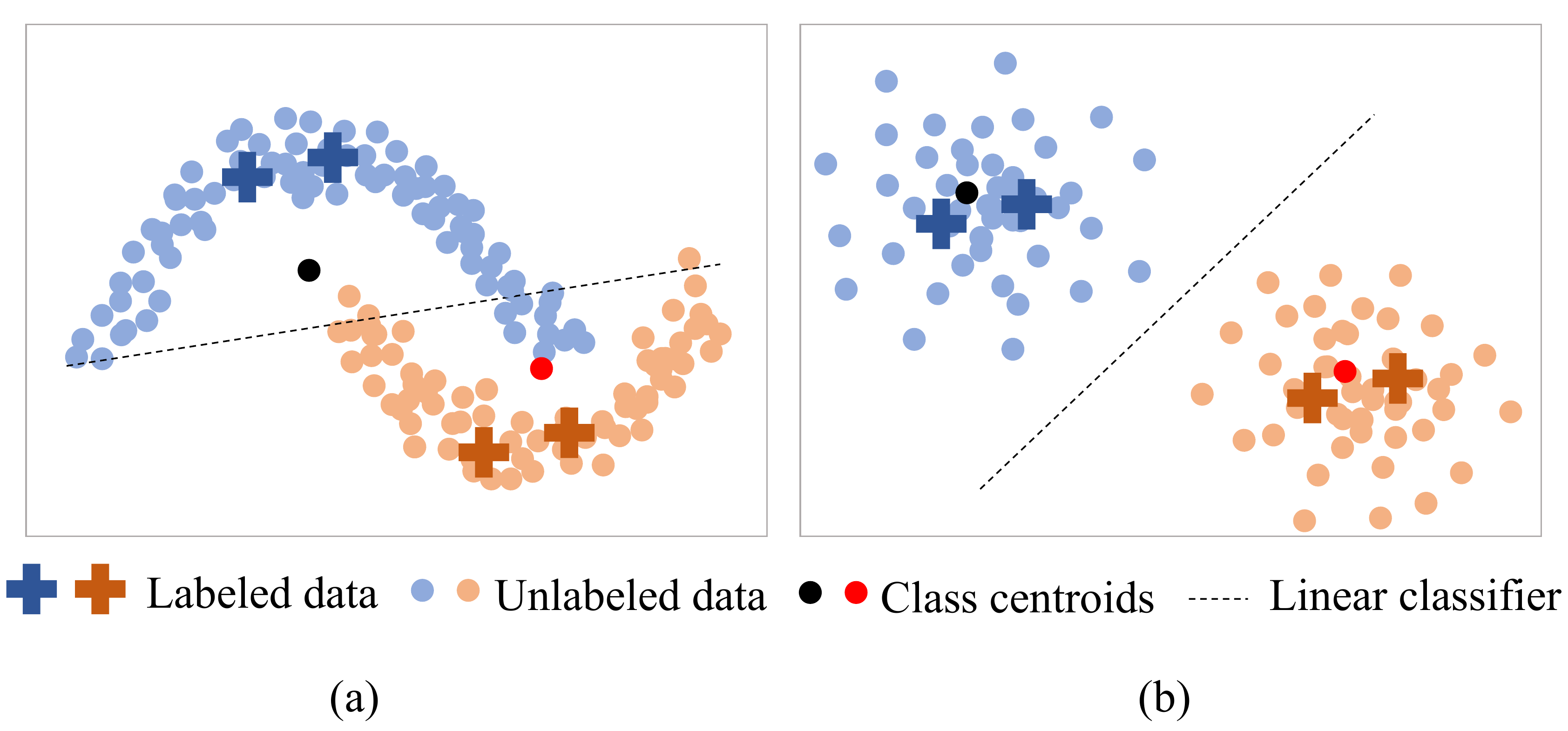}
\caption{
Two toy examples which (a) visualizes the feature space of an indiscriminative semi-supervised model, and (b) visualizes the feature space of a well-clustered semi-supervised model. 
}
\label{fig:toy_example}
\end{figure}

A common assumption of semi-supervised learning is that the decision boundary should not pass through high-density regions. 
Consistency regularization-based techniques~\cite{DBLP:conf/miccai/YuWLFH19,luo2021urpc,li2020transformation} achieve a decision boundary at a low-density area by penalizing prediction variation under different input perturbations.
Entropy minimization-based methods aim to achieve high-confidence predictions for unlabeled data either in an explicit manner~\cite{DBLP:conf/nips/GrandvaletB04} or an implicit manner~\cite{lee2013pseudo,DBLP:conf/miccai/SedaiARJ0SWG19,DBLP:journals/tmi/FanZJZCFSS20,reiss2021every}.
As shown in Figure~\ref{fig:toy_example}, an ideal model should pull together data points of the same class and push apart data points from different classes in the feature space. 
As the training set of semi-supervised learning includes labeled and unlabeled images, it is challenging to directly optimize the unlabeled images in the feature space without explicit guidance. 
We observe that with unlabeled images, most semi-supervised methods~\cite{DBLP:conf/miccai/YuWLFH19,luo2021urpc,li2020transformation} can achieve more accurate segmentation results than the model trained with only labeled data. 
Therefore, the pseudo segmentation predicted by a semi-supervised model on unlabeled data could possibly be made even more stable and precise.

Motivated by this observation, we present a simple yet effective two-stage framework for semi-supervised medical image segmentation with the key idea to \emph{explore representation learning for segmentation from both labeled and unlabeled images}. 
The first stage aims to generate high-quality pseudo labels, and the second stage aims to use pseudo labels to retrain the network to regularize features for both labeled and unlabeled images.  
Existing uncertainty-based semi-supervised methods~\cite{DBLP:conf/miccai/YuWLFH19,DBLP:journals/tmi/CaoCLPWC21,DBLP:conf/miccai/WangZTZSZH20,wang2022semi} have achieved stunning results by considering the reliability of the supervision for the unlabeled images.
These methods exploit the epistemic uncertainty, a kind of uncertainty about the model's parameters arising from a lack of data, either in the output space~\cite{DBLP:conf/miccai/YuWLFH19,DBLP:journals/tmi/CaoCLPWC21,DBLP:conf/miccai/WangZTZSZH20} or in the feature space~\cite{DBLP:conf/miccai/WangZTZSZH20}, as guidance for identifying trustworthy supervision.
Medical images are often noisy, and the boundaries between tissue types may not be well defined, leading to a disagreement among human experts~\cite{DBLP:conf/nips/KohlRMFLMERR18, DBLP:conf/miccai/BaumgartnerTCHM19, DBLP:conf/nips/MonteiroFCPMKWG20}. 
However, aleatoric uncertainty that represents the ambiguity about the input data and is irreducible by obtaining more data, is ignored in these methods.

To obtain high-quality pseudo labels for unlabeled images, we present an Aleatoric Uncertainty Adaptive method, namely AUA, for semi-supervised medical image segmentation. 
Under the framework of the mean teacher model~\cite{tarvainen2017mean}, to obtain reliable target supervision for unlabeled data, instead of estimating the model's epistemic uncertainty~\cite{DBLP:conf/miccai/YuWLFH19,DBLP:journals/tmi/CaoCLPWC21,DBLP:conf/miccai/WangZTZSZH20}, we explore the aleatoric uncertainty of the model for noisy input data.  
AUA first measures the spatially correlated aleatoric uncertainty by modeling a multivariate normal distribution over the logit space. 
To effectively utilize unlabeled images, AUA encourages the prediction consistency between the teacher model and the student model by adaptively considering the aleatoric uncertainty for each image. 
Specifically, the consistency regularization automatically emphasizes the input images with lower aleatoric uncertainty, \ie, input images with less ambiguity.

In the second stage, we retrain the network with pseudo labels. 
To effectively regularize feature representation learning in both stages, we propose stage-adaptive feature regularization, including a \emph{boundary-aware contrastive loss} in the first stage and a \emph{prototype-aware contrastive loss} in the second stage. 
The main idea of boundary-aware contrastive loss is to \textit{fully leverage labeled images for representation learning.} 
A straightforward solution is to pull together the pixels to the same class and push away pixels from different classes using a contrastive loss. 
However, medical images usually contain a large number of pixels. Simply utilizing contrastive loss would lead to a high computational cost and memory consumption. 
To this end, we present a boundary-aware contrastive loss, where only randomly sampled pixels from the segmentation boundary are optimized. 
In the second stage, to \emph{effectively utilize both labeled and pseudo-labeled images}, \ie, \emph{unlabeled images for representation learning},
we present a prototype-aware contrastive loss with each pixel's feature 
pulled closer to its class centroid, \ie, prototype, and pushed further away from the class centroids it does not belong to.
The main intuition is that the trained model can \emph{generate pseudo labels for unlabeled images in the second stage}. 
Compared with the boundary-aware contrastive loss, the prototype-aware contrastive loss better leverages the pseudo labels, especially those that may not occur at the segmentation boundaries. 

In summary, this paper makes the following contributions: 
\begin{itemize}
    \reviseagain{
    \item 
    We introduce stage-adaptive contrastive losses (\ie, boundary-aware contrastive loss and prototype-aware contrastive loss) to regularize a more compact and better-separated feature space, which eases the learning of a segmentation decision boundary.
    
    \item  We present AUA, an aleatoric uncertainty adaptive consistency regularization method that paves the way for prototype-aware contrastive loss by improving pseudo label quality and prototype estimation.

    \item Our method achieves state-of-the-art performance on three public datasets. The ablation study validates the effectiveness of our proposed method. Our code is available at GitHub~\url{https://github.com/Huiimin5/AUA} now.
}
\end{itemize}

\section{Related Work}



We briefly discuss related works in semi-supervised medical image segmentation, including pseudo labeling and consistency regularization. We also discuss some techniques related to contrastive learning and uncertainty estimation.


\subsection{Semi-supervised Medical Image Segmentation}
Semi-supervised learning (SSL) refers to training the model with both labeled and unlabeled images. 
\revise{
A wide span of tasks has been explored such as segmentation~\cite{li2020transformation}, classification~\cite{6153383,9992211,9927317,9738732}, and crowd counting~\cite{10040995}.
}
For medical image segmentation, 
Early work used graph-based methods~\cite{DBLP:journals/tmi/SuYHKZ16,DBLP:conf/icpr/BorgaAL16} for semi-supervised segmentation.
Recently, semi-supervised medical image segmentation has featured deep learning. The existing methods can be broadly classified into two categories: pseudo labeling-based~\cite{lee2013pseudo,reiss2021every,DBLP:conf/cvpr/XieLHL20,DBLP:journals/corr/abs-2012-00827,DBLP:journals/mia/XiaYYLCYZXYR20} and consistency regularization-based methods~\cite{DBLP:conf/miccai/YuWLFH19,luo2021urpc,li2020transformation,DBLP:journals/tmi/CaoCLPWC21,DBLP:conf/miccai/WangZTZSZH20,DBLP:conf/miccai/HangFLYWCQ20,DBLP:conf/miccai/FangL20, DBLP:journals/corr/abs-2103-02911,luo2021semi,li2018semi,  DBLP:conf/miccai/BortsovaDHKB19,DBLP:conf/miccai/LiZH20,DBLP:conf/miccai/YangSKW20}. 

\noindent \textbf{Pseudo Labeling-based Methods.} 
Pseudo labeling-based methods handle label scarcity by estimating pseudo labels on unlabeled data and using all the labeled and pseudo labeled data to train the model.
Self-training is one of the most straightforward solutions ~\cite{lee2013pseudo,DBLP:conf/cvpr/XieLHL20,DBLP:journals/corr/abs-2012-00827} and has been extended to the biomedical domain for segmentation~\cite{DBLP:conf/miccai/SedaiARJ0SWG19,DBLP:journals/tmi/FanZJZCFSS20,DBLP:conf/miccai/BaiOSSRTGKMR17}.
The main idea of self-training is that the model is first trained with labeled data only and then generates pseudo labels for unlabeled data. By retraining the model with both labeled and pseudo labeled images, the model performance can be enhanced.  
The model can be trained iteratively with these two processes until the model  performance becomes stable and satisfactory. 
To reduce the noise in pseudo labels, different methods have been developed, including identifying trustworthy pseudo labels by uncertainty estimation~\cite{DBLP:conf/miccai/SedaiARJ0SWG19}, using a 
Conditional Random Field (CRF)~\cite{DBLP:conf/nips/KrahenbuhlK11} to refine pseudo labels~\cite{DBLP:conf/miccai/BaiOSSRTGKMR17} or using pseudo labels only for fine-tuning~\cite{DBLP:journals/tmi/FanZJZCFSS20}.
In addition to such an offline pseudo label generation strategy, online self-training methods~\revise{\cite{DBLP:conf/cvpr/VuJBCP19, DBLP:conf/miccai/LiCXMZ20,reiss2021every}} have been developed recently where pseudo labels are generated after each forward propagation and used as immediate supervision.

Another pseudo labeling-based method is co-training ~\cite{DBLP:conf/colt/BlumM98,DBLP:conf/eccv/QiaoSZWY18,DBLP:conf/ijcai/ChenWGZ18} where multiple learners are trained and their disagreement on unlabeled data is exploited for improving the accuracy of pseudo labels. 
The basic idea is that each learner could learn different and complementary information from the other learners. 
In some self-training methods, more than one learner can be used, such as in~\cite{reiss2021every} and the supervision on unlabeled data is unidirectional. For example, the teacher model~\cite{tarvainen2017mean} generates pseudo labels to supervise the student model, while in a dual-model co-training method such as~\cite{DBLP:journals/mia/XiaYYLCYZXYR20}, supervision is bidirectional. Specifically, each base model's supervision of unlabeled data is based on the fused predictions from the other base models, weighted by the confidence of each model.

However, these methods ignore the class-aware feature regularization,
which is a key focus of this study. We will demonstrate the importance of feature representation learning in learning with labeled and pseudo labeled images. 

\noindent \textbf{Consistency regularization-based Methods.} 
The goal of consistency regularization-based semi-supervised methods~\cite{tarvainen2017mean,DBLP:conf/iclr/LaineA17,DBLP:journals/pami/MiyatoMKI19} is to find the model that is not only accurate in predictions but also invariant to input perturbations to enforce the decision boundary traverse the low-density region of the feature space.
One line of these methods considers invariance to input domain perturbations. For example, the temporal ensembling model~\cite{DBLP:conf/iclr/LaineA17} achieves promising results by accumulating soft pseudo labels on randomly perturbated input images.
An extension work with soft pseudo label accumulation guided by epistemic uncertainty was proposed in~\cite{DBLP:journals/tmi/CaoCLPWC21}.
When the epistemic uncertainty of the prediction is high, it will contribute less to pseudo label accumulation. 
The mean teacher model~\cite{tarvainen2017mean} achieves invariance to input perturbations by promoting consistency between the predictions of the teacher and the student models where input images fed to the teacher model are added with noises.
Extensions have also been made from the perspective of reliability evaluation\reviseagain{~\cite{DBLP:conf/miccai/YuWLFH19,DBLP:conf/miccai/WangZTZSZH20,wang2022semi}} to provide reliable supervision from the teacher model to the student model or considering structural information of foreground objects\reviseagain{~\cite{DBLP:conf/miccai/HangFLYWCQ20,wang2022semi,you2022simcvd}}.
In addition to input domain perturbation, other perturbations that would not change the semantics of the prediction have also been designed and used to promote consistency. 
\revise{
For example, consistency among predictions given by differently designed decoders~\cite{DBLP:conf/miccai/FangL20, DBLP:journals/corr/abs-2103-02911,wu2022mutual,DBLP:conf/cvpr/OualiHT20}, or at different scales~\cite{luo2021urpc} or with different modalities~\cite{luo2021semi} where perturbations are beyond the input level is maintained. 
The distribution-level consistency between predicted segmentation maps on labeled data with those on unlabeled ones~\cite{DBLP:conf/miccai/LiZH20,DBLP:conf/miccai/ZhangYCFHC17} has also been proved effective.
}
Aside from perturbations that lead to invariance in output, there is another line of studies~\cite{li2018semi, li2020transformation, DBLP:conf/miccai/BortsovaDHKB19} that promotes equivariance between the input and the output because some input space transform, especially spatial transform such as rotations, should lead to the same transform in the output space.


Unlike these existing methods that are based on consistency regularization, our method is a two-stage framework, which improves the overall framework by regularizing the feature representation. 
Moreover, we introduce AUA, an aleatoric uncertainty-aware method, to represent inherent ambiguities in medical images and enhance the segmentation performance by encouraging consistency for images with low ambiguity.



\subsection{Contrastive Learning in Semi-supervised Image Segmentation.}

Note that we exclude self-supervised learning methods where unlabeled data are only used for task-agnostic purposes, \ie, pretraining 
such as in~\cite{DBLP:conf/nips/ChaitanyaEKK20}, even though performance under semi-supervised setting is also reported.
We only consider contrastive learning for task-specific use~\cite{DBLP:conf/brainles-ws/IwasawaHS20,lai2021semi,DBLP:journals/corr/abs-2012-06985}.
Among these works, only~\cite{DBLP:journals/corr/abs-2012-06985}'s goal is to promote inter-class separation and intra-class compactness.
However,  
in~\cite{DBLP:journals/corr/abs-2012-06985}, pseudo labels are obtained from the model trained with labeled data only, whose performance is inferior to our first stage model, where pseudo labels are obtained from a model that takes advantage of consistency regularization on unlabeled data and feature regularization on labeled data.
In~\cite{lai2021semi}, inter-class separation is considered by taking pixels with different pseudo labels as negative pairs, but intra-class compactness is ignored since the positive pair is built on the same pixels from different crops, which essentially is an extension of instance discrimination for the segmentation task.
To the best of our knowledge, 
ours is the first study with pixel-level feature regularization aiming at intra-class compactness and inter-class separation
for semi-supervised medical image segmentation.

\subsection{Uncertainty Estimation in Semi-supervised Medical Image Segmentation.}
Uncertainties generally fall into two categories: epistemic and aleatoric.
Epistemic uncertainty is about a model's parameters caused by a lack of data while aleatoric uncertainty is caused by intrinsic ambiguities or randomness of input data and cannot be reduced by introducing more data.
Early methods measure uncertainty using particle filtering and Conditional Random Fields 
(CRFs)~\cite{DBLP:journals/ijcv/BlakeCZ93,DBLP:conf/cvpr/HeZC04}.
More recently, in Bayesian networks, epistemic uncertainty is usually estimated with Monte Carlo Dropout~\cite{DBLP:conf/nips/KendallG17}, which has been extended for the semi-supervised medical image segmentation task~\cite{DBLP:conf/miccai/YuWLFH19,DBLP:journals/tmi/CaoCLPWC21,DBLP:conf/miccai/WangZTZSZH20}.
Aleatoric uncertainty is estimated either without considering correlations between pixels~\cite{DBLP:conf/nips/KendallG17} or 
with a limited ability to model spatial correlation since it is captured by uncorrelated latent variables from multivariate normal distribution~\cite{DBLP:conf/nips/KohlRMFLMERR18,DBLP:conf/miccai/BaumgartnerTCHM19}.
Monteiro et al.~\cite{DBLP:conf/nips/MonteiroFCPMKWG20} proposed an aleatoric uncertainty estimation technique where correlations between pixels are considered.
Given the ubiquitous existence of noises or ambiguities in medical images, aleatoric uncertainty has been overlooked for semi-supervised medical image segmentation.
In this work, we propose an aleatoric uncertainty adaptive consistency regularization technique, where correlations between pixels are considered when measuring aleatoric uncertainty.

\section{Method}
\label{sec:method}
\begin{figure*}[ht]
    \centering
    \includegraphics[width=0.95\textwidth]{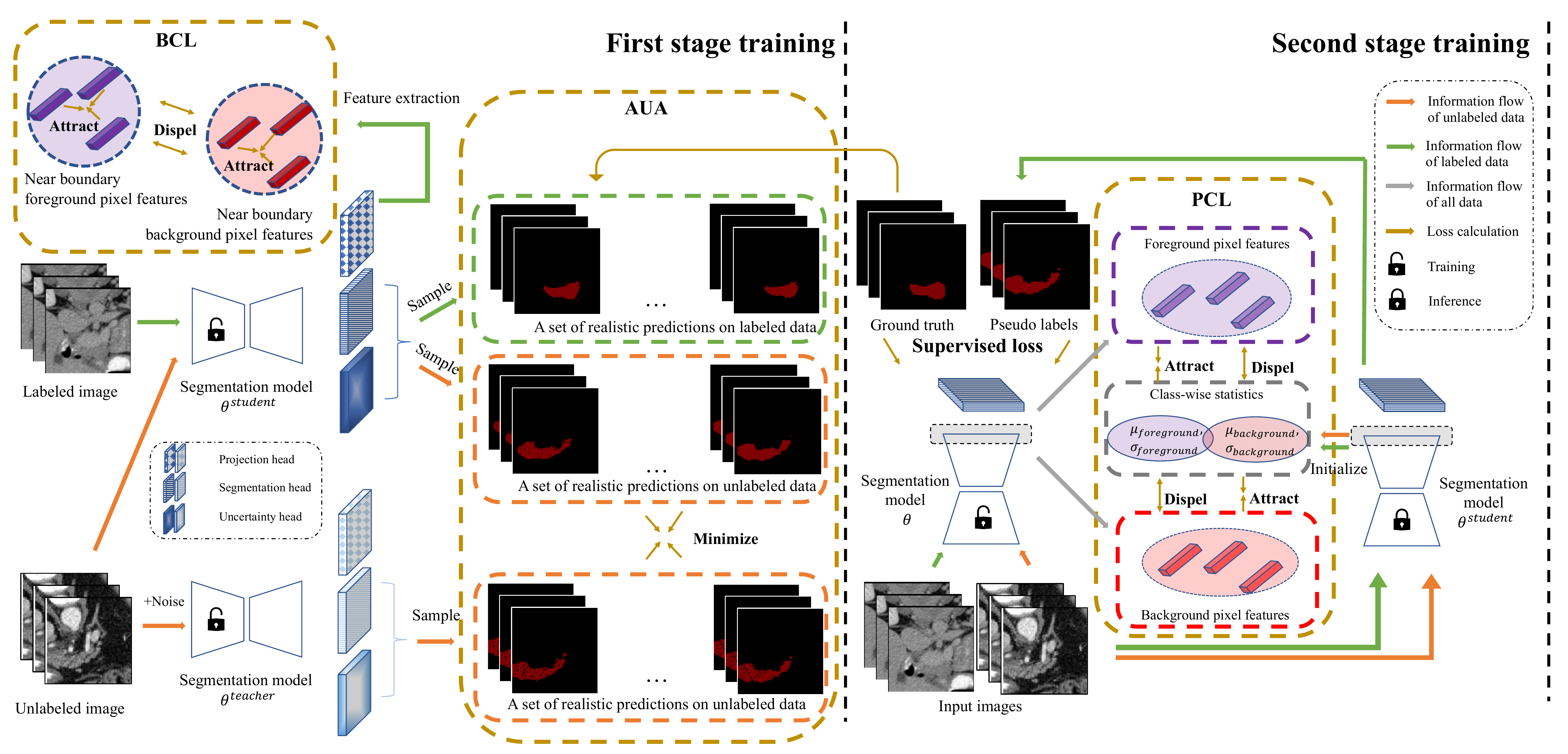}
    \caption{Overview of our method.
    \reviseagain{The proposed loss functions (AUA, BCL and PCL) are boxed out in \textcolor[rgb]{0.7,0.4,0.2}{brown}. Firstly, we propose AUA under a mean-teacher framework which consists of a student model (parameterized by $\theta^{student}$) 
    and a teacher model (parameterized by 
    $\theta^{teacher}$).
    AUA is composed of an aleatoric uncertainty estimation module (boxed out in \textcolor[rgb]{0,0.8,0.1}{green}) and an adaptive consistency regularization module (boxed out in \textcolor[rgb]{1.0,0.43,0.0}{orange}).
    Secondly, stage-wise contrastive learning is proposed, which consists of BCL on boundary pixels of labeled data in the first stage, as well as a prototype-aware contrastive loss which is applied to all pixels in the second stage.}
    }
    \label{fig:overview}
\end{figure*}

Figure~\ref{fig:overview} visualizes the overview of the proposed two-stage framework. The input image is first fed into the AUA module to get a segmentation model which generates a high-quality pseudo label. 
Then, we introduce the stage-adaptive contrastive learning method, consisting of a boundary-aware contrastive loss (BCL) on labeled data only in the first stage and a prototype-aware contrastive loss (PCL) on all data in the second stage. By sequential training through the first and second stages, we generate the final segmentation results. 

\revise{
\subsection{Preliminaries: Deterministic Medical Image Segmentation}}
Here, we consider a 
$C$-class
segmentation task on 3D volumes with size $H\times W\times D\times C$, where 
$H$, $W$, and $D$ denote the height, width, and depth, respectively.
Given an image $x \in R^{H\times W\times D\times C}$ and its ground truth $y$
with the same size,
the loss function of a general segmentation network is designed to minimize the negative log-likelihood, formulated as:
\begin{align}
\label{equ:sup_general}
\mathcal{L}_{sup} =- \log p(y|x)=-\log\int p(y|g)p(g|x)dg,
\end{align}
where $g$
denotes logits.

In a deterministic segmentation network, \ie, assuming 
$p(g|x)=\delta(f(g|x;\theta))$ and independence of each pixel's prediction on the other, where 
$f$ is
a neural network parameterized by 
$\theta$
and $\delta$
denotes the Dirac delta function,
the loss function in Eq.~\ref{equ:sup_general} can be rewritten as:

\begin{small}
\begin{align}
\label{equ:sup_ce}
\mathcal{L}_{sup_{ce}} =-\log p(y|g)=-\sum_{i=1}^{V}\sum_{c=1}^Cy_{ic}{\rm \log  softmax}(g_i)_c.
\end{align}
\end{small}
For simplicity, we use a one-dimensional scalar $i$
to index each pixel out of a whole set of $V=H*W*D$ 
pixels in a 3D volume.
The above equation is the cross-entropy function commonly used in segmentation models.

\revise{
\subsection{Preliminaries: Aleatoric Uncertainty Estimation for Segmentation}}
To model inherent ambiguities of input data,
we follow~\cite{DBLP:conf/nips/MonteiroFCPMKWG20} and assume a multi-variant Gaussian distribution around logits, \ie, $g|x\sim N(\mu(x), \sigma(x))$, with $\mu(x)\in R^{H\times W\times D\times C}$ and $\sigma(x)\in R^{{(H\times W\times D\times C)}^2}$.
Monte-Carlo integration of $S$ samples is applied to approximate the intractable integral operation,
leading us from Eq.~\ref{equ:sup_general} to:
\begin{align}
\label{equ:sup_au}
\mathcal{L}_{sup_{au}} &=-\log\frac{1}{S}\sum_{s=1}^{S}p(y|g^{(s)})\\
&=-{\rm logsumexp}_{s=1}^{S}\log p(y|g^{(s)})+\log(S).
\end{align}
\reviseagain{
The logsumexp (LSE)
 operation is defined as LSE($l_1,...,l_S$)=
 $log(exp(l_1)+...+exp(l_S))$
 where $l_s=\log p(y|g^{(s)})
 $.
}
We refer the calculation of 
$\log p(y|g^{(s)})$
to Eq.~\ref{equ:sup_ce},
where
$g^{(s)}$ 
is a sample out of 
$g|x\sim N(\mu(x), \sigma(x))$.
As pointed out by \cite{DBLP:conf/nips/MonteiroFCPMKWG20},
the full-rank covariance matrix $\sigma(x)$ is computationally infeasible, so we also adopt a low-rank (specifically, $r$-rank) approximation defined as:
\begin{equation}
\label{equ:stage1_cov}
    \sigma(x)=\widetilde{F}\widetilde{F}^T+\widetilde{D},
\end{equation}
where $\widetilde{F} \in R^{H\times W\times D\times C \times r}$
denotes the factor part of a low-rank form of the covariance matrix and $\widetilde{D} \in R^{H\times W\times D\times C}$ denotes the diagonal part.
Compared with a full-rank parameterization, where $(H\times W\times D\times C)^2$ parameters should be estimated, which is beyond what a GPU card can accommodate unless for very small images, 
its low-rank approximation is more computationally feasible since the complexity reduces from quadratic to linear, \ie, 
$H\times W\times D\times C \times r $ (for $\widetilde{F}$)
$ + H\times W\times D\times C$ (for $\widetilde{D}$).
This approximation might lead to a compromise of estimated aleatoric uncertainty but still can bring effective guidance for semi-supervised learning, as shown in our experiments.

\subsection{Aleatoric Uncertainty Adaptive Consistency Regularization (AUA)}
\label{stage1:consist}
\revise{It is desirable if the semi-supervised segmentation model can be aware of its chance of making mistakes on unlabeled data.
We resort to aleatoric uncertainty that captures input ambiguities,
to guide how much the student model should learn from the teacher model.
A consistency regularization technique adaptive to aleatoric uncertainty is proposed.
}
Given an unlabeled image $x^u$, 
the predicted distribution by the student model parameterized by $\theta^{s}$ is denoted as
$p_s= p(y^u|x^u; \theta^{s})$.
Similarly, we can obtain the teacher model's prediction 
$p_t= p(y^u|{x^u}'; \theta^{t})$
over the perturbed version of the same input ${x^u}'$ 
by Gaussian noise injection, 
where parameters of the teacher model, denoted as $\theta^{t}$, are updated with an exponential moving average of the parameters of the student model.
The consistency between the teacher model's predictions and the student model's predictions on unlabeled data is encouraged by minimizing the generalized energy distance \cite{DBLP:conf/nips/KohlRMFLMERR18, szekely2013energy}, which is defined as:
\begin{equation}
\begin{aligned}
\label{equ:stage1_ged_orig}
\mathcal{L}_{con}^{'}&=2E_{y_s\sim p_s, y_t\sim p_t}d(y_s, y_t)\\&-E_{y_{s1}\sim p_s, y_{s2}\sim p_s}d(y_{s1},y_{s2})\\&-E_{y_{t1}\sim p_t, y_{t2}\sim p_t}d(y_{t1},y_{t2}).
\end{aligned}
\end{equation}
To approximate the intractable expectation operation in Eq.~\ref{equ:stage1_ged_orig}, 
we take $S$
samples out of $p_s$ and $p_t$, respectively. 
The consistency regularization loss function can be reformulated as:
\begin{equation}
\begin{aligned}
\label{equ:stage1_ged}
\mathcal{L}_{con}&=2\sum\nolimits_{i_s=0}^{S}\sum\nolimits_{i_t=0}^{S}d(y_s^{(i_s)}, y_t^{(i_t)})
\\&-\sum\nolimits_{i_{s1}=0}^{S}\sum\nolimits_{i_{s2}=0}^{S}d(y_s^{(i_{s1})}, y_s^{(i_{s2})})
\\&-\sum\nolimits_{i_{t1}=0}^{S}\sum\nolimits_{i_{t2}=0}^{S}d(y_t^{(i_{t1})}, y_t^{(i_{t2})}),\\
&y_s^{(i_s)}, y_s^{(i_{s1})}, y_s^{(i_{s2})}\sim p_s,\\
&y_t^{(i_t)}, y_t^{(i_{t1})}, y_t^{(i_{t2})}\sim p_t.\\
\end{aligned}
\end{equation}
In Eq.~\ref{equ:stage1_ged},
$d$
is defined as the Generalized Dice loss~\cite{sudre2017generalised}:
\begin{equation}
\begin{aligned}
&d(y^i, y^j)=1-\\ &\frac{\sum_{k=1}^{H\times W\times D}\sum_{c=1}^{C}(y_{kc}^i\cdot y_{kc}^j)}{\sum_{k=1}^{H\times W\times D}\sum_{c=1}^{C}(y_{kc}^i\cdot y_{kc}^i)+\sum_{k=1}^{H\times W\times D}\sum_{c=1}^{C}(y_{kc}^j\cdot y_{kc}^j)},
\end{aligned}
\end{equation}
where $k$
indexes each pixel out of a whole set of $V=H*W*D$ 
pixels in a 3D volume
and $c$
indexes each class out of a total of $C$ classes.

The optimum of Eq.~\ref{equ:stage1_ged} is 0, which means the optimum of the first term is the sum of the last two.
This consistency regularization metric is adaptive to aleatoric uncertainty in the sense that if the diversity between samples of the student (or the teacher) model is high, \ie, the values of the last two terms of Eq.~\ref{equ:stage1_ged} are large, indicating a high aleatoric uncertainty, the pairwise similarity of samples from the student and the teacher models, denoted by the first term of Eq.~\ref{equ:stage1_ged}, would be less strictly constrained.
On the contrary, on input data where a low diversity is estimated, implying the aleatoric uncertainty is low and the model is more likely to generalize well, 
the student model automatically learns more from the teacher model by optimizing the first term to a smaller value.

To summarize,  AUA loss is defined as follows:
\begin{align}
\label{equ:aua}
\mathcal{L}_{AUA} = \mathcal{L}_{sup_{au}} + \lambda_g\mathcal{L}_{con},
\end{align}
where $\lambda_g$
is the scaling weight to balance the uncertainty estimation loss and the generalized energy distance loss.

\subsection{Stage-adaptive feature regularization}
\label{contrast}
We introduce a stage-adaptive feature learning method consisting of a boundary-aware contrastive loss and a prototype-aware contrastive loss, to enhance the representation learning with \emph{only labeled images} and \emph{both labeled and pseudo labeled images}, respectively. 
A natural solution is a contrastive loss with features of pixels belonging to the same class (\ie, both foreground pixels or both background pixels) as positive pairs and features belonging to different classes (\ie, one from foreground the other from background) as negative pairs. 
This strategy allows us to perform pixel-wise regularization but consumes memory quadratically to the number of pixels,
so we propose a stage-adaptive contrastive learning method with these concerns properly handled.
To reduce the computational cost, at the first stage, we only optimize the feature representation for pixels around the segmentation boundaries, using a boundary-aware contrastive loss (BCL). 
At the second stage, with more accurate pseudo labels on unlabeled data, we introduce a prototype-aware contrastive loss (PCL) to fully leverage both labeled and pseudo labeled images for representation learning. 


\subsubsection{Boundary-aware contrastive learning}
\label{stage1:sup_contrast}


As a balance of benefits of pixel-wise feature level regularization and computational costs, 
we build positive and negative pairs based on a random subset of near-boundary pixels, arriving at the 
boundary-aware contrastive loss formally defined as:
\begin{align}
\label{equ:sup_contr}
\mathcal{L}_{BCL}&=\sum_{i\in NB}\frac{-1}{P(i)}\sum_{pi\in P(i)}
\log\frac{exp(f^1_i\cdot f^1_{pi}/ \tau_1)}{\sum\limits_{o\in O(i)}exp(f^1_i\cdot f^1_o/\tau_1)},
\end{align}
where $NB$ contains indexes of randomly sampled near boundary pixels from an input image, $O(i)$ contains indexes of the other pixels except pixel $i$
and $P(i)$
contains indexes of pixels in $O(i)$
belonging to the same class as pixel $i$.
The feature vectors $f^1_i$, $f^1_o$ and $f^1_{pi}$
are obtained from a 3-layer convolutional projection head, which is connected after the layer before the last layer.
The temperature $\tau_1$ is set to be 0.07. 
\revise{
By sub-sampling, BCL reduces the computational cost from $(H\times W\times D)^2$
(i.e., pixel-wise contrastive loss) to  $NB^2$.
}

\subsubsection{Prototype-aware contrastive learning}
\label{stage2:sup_contrast}
In the second stage,
the way to regularize an indiscriminative feature space as in Figure~\ref{fig:toy_example}(a) is to encourage each feature to be closer to any other pixels that share the same label and further away from the centroid of opposite class so that forming a feature space in Figure~\ref{fig:toy_example}(b) is encouraged, which is defined as:
\begin{equation}
\begin{aligned}
\label{equ:stage2_contrastive_orig}
&\mathcal{L}_{PCL}'=-\frac{1}{|\mathcal{P}|}\sum_{i\in \mathcal{P}}\frac{-1}{|P(i)\backslash \{i\}|}
\\&\sum_{pi\in P(i)}\log\frac{exp(f_i^2\cdot f_{pi}^2/ \tau)}{exp(f_i^2\cdot f_{pi}^2/ \tau)+\frac{1}{|N(i)|}\sum_{ni\in N(i)}exp(f_i\cdot f_{ni}^2/\tau)},
\end{aligned}
\end{equation}
where $\mathcal{P}$ contains indexes of all pixels.
$P(i)$ and $N(i)$ 
contains the indexes of positive pixels, \ie, those sharing the same class,
and negative pixels, \ie, those with different labels
to pixel $i$ 
, respectively.
Features extracted from the second stage model are denoted as $f_*^2$
where $*$
can be an index of any pixel.

In~\cite{DBLP:journals/corr/abs-2105-05013}, by assuming a Gaussian distribution for features belonging to each class, the computational cost of Eq.~\ref{equ:stage2_contrastive_orig} can be reduced from quadratic to linear, leading to a regularization formulated as:

\begin{equation}
\begin{aligned}
\label{equ:stage2_contrastive}
&\mathcal{L}_{PCL}={f^2_i}^\top\sigma_p f^2_i-
\\&\frac{1}{|\mathcal{P}|}
\sum_{i\in \mathcal{P}}\log\frac{exp(\frac{f^2_i\cdot \mu_p}{\tau_2}+\frac{{f^2_i}^\top\sigma_p f^2_i}{2\tau_2^2})}{exp(\frac{f^2_i\cdot \mu_p}{\tau_2}+\frac{{f^2_i}^\top\sigma_p f^2_i}{2\tau_2^2})+exp(\frac{f^2_i\cdot \mu_n}{\tau_2}+\frac{{f^2_i}^\top\sigma_n f^2_i}{2\tau_2^2})},
\end{aligned}
\end{equation}
where $\mu_p$ and $\sigma_p$ are the mean and covariance matrix of the positive class to pixel $i$ and similarly, $\mu_n$ 
and $\sigma_n$ are the mean and covariance matrix of the negative class corresponding to pixel $i$.
These prototype statistics for each class $c$
are estimated from the first stage model with an
moving average update of extracted features
with each update at $t$-step
formulated as:

\begin{footnotesize}
\begin{equation}
\begin{aligned}
\label{equ:statistics}
\mu^c_t&=\frac{N_{t-1}^c\mu^c_{t-1}+n_{t}^c\mu^{'c}_t}{N_{t-1}^c+n_{t}^c},\\
\sigma^c_t&=\frac{N_{t-1}^c\sigma^c_{t-1}+n_{t}^c\sigma^{'c}_t}{N_{t-1}^c+n_{t}^c}
+\frac{N_{t-1}^cn_{t}^c(\mu^c_{t-1}-\mu^{'c}_{t})(\mu^c_{t-1}-\mu^{'c}_{t})^\top}{(N_{t-1}^c+n_{t}^c)^2},
\end{aligned}
\end{equation}
\end{footnotesize}
where $N_{t-1}$ 
denotes the total number of pixels belonging to class $c$
seen before time step $t$,
and $n_t$
denotes the pixel number of class $c$ 
in the loaded image at time step $t$.
$\mu^{'c}_t$ and $\sigma^{'c}_t$ 
denote the mean and covariance, respectively, of features belonging to class $c$
in images at $t$.
The final prototypes are estimated after 3000 iterations and the temperature $\tau_2$ is set to be 100.
\revise{By utilizing prototypes, BCL reduces the computational cost from $(H\times W\times D)^2$
(i.e., pixel-wise contrastive loss) to  $H\times W\times D \times C$.
}


\subsection{Stage-wise Training as a Unified Framework}
\label{method:overall}
To summarize, in the first stage, the loss function is defined as:
\begin{align}
\label{equ:stage1}
\mathcal{L}_{stage1} = \mathcal{L}_{AUA} + \lambda_c\mathcal{L}_{BCL}
\end{align}
where $\lambda_c$ is the scaling weight for BCL loss.
To this end, pseudo labels on unlabeled data with higher quality can be obtained thanks to joint prediction regularization (with AUA) and feature regularization (with BCL),
which enables retaining a stronger segmentation model at the second stage by regularizing both predictions and features over the whole dataset in a label-aware manner.
The loss function in the second stage is as follows:
\begin{align}
\mathcal{L}_{stage2} = \mathcal{L}_{sup_{ced}} + \lambda_r\mathcal{L}_{PCL}
\end{align}
where $\mathcal{L}_{sup_{ced}}$
is defined as the average of cross-entropy loss and Dice loss as a common practice in segmentation, which serves as pseudo labeling
and $\lambda_r$
is the weight for PCL loss.

\begin{table*}[ht]
\centering
\caption{A comparison with state-of-the-art on pancreas dataset with 20\% labeled data. The up arrow implies that the larger the number, the better the performance.
The down arrow implies that a lower number indicates a better performance.
}\label{tab:results_pancreas_cmp}
\begin{tabular}{c|c|c|c|c|c|c}
\toprule[1.5pt]
\multirow{2}{*}{Method} & \multicolumn{2}{c}{\#\# scans used} & \multicolumn{4}{|c}{Metrics}\\
\cline{2-7}
&Labeled&Unlabeled&Dice[\%]$\uparrow$&Jaccard[\%]$\uparrow$&ASD[voxel]$\downarrow$&95HD[voxel]$\downarrow$\\
\hline
V-Net&12&0&70.63&56.72&6.29&22.54\\
V-Net&62&0&81.78&69.65&1.34&5.13\\
\hline
MT~\cite{tarvainen2017mean}&12&50&75.85&61.98&3.40&12.59\\
DAN~\cite{DBLP:conf/miccai/ZhangYCFHC17}&12&50&76.74&63.29&2.97&11.13\\
Entropy Mini~\cite{DBLP:conf/cvpr/VuJBCP19}&12&50&75.31&61.73&3.88&11.72\\
UA-MT~\cite{DBLP:conf/miccai/YuWLFH19}&12&50&77.26&63.82&3.06&11.90\\
CCT~\cite{DBLP:conf/cvpr/OualiHT20}&12&50&76.58&62.76&3.69&12.92\\
SASSNet~\cite{DBLP:conf/miccai/LiZH20}&12&50&77.66&64.08&3.05&10.93\\
DTC~\cite{luo2021semi}&12&50&78.27&64.75&2.25&8.36\\
\reviseagain{
URPC~\cite{luo2021urpc}}&12&48&78.83&66.01&1.96&6.92\\
\reviseagain{
MC-Net~\cite{wu2022mutual}}&12&48&79.27&66.24&1.90&7.07\\
\hline
Ours&12&48&\textbf{79.81}&\textbf{66.82}&\textbf{1.64}&\textbf{5.90}\\
\bottomrule[1.5pt]
\end{tabular}
\end{table*}

\begin{table*}[!ht]
\centering
\caption{A comparison with state-of-the-art on Pancreas dataset with 5\% labeled data.
}\label{tab:low_pancreas}
\begin{tabular}{c|c|c|c|c|c|c}
\toprule[1.5pt]
\multirow{2}{*}{Method} & \multicolumn{2}{c}{\#\# scans used} & \multicolumn{4}{|c}{Metrics}\\
\cline{2-7}
&Labeled&Unlabeled&Dice[\%]$\uparrow$&Jaccard[\%]$\uparrow$&ASD[voxel]$\downarrow$&95HD[voxel]$\downarrow$\\
\hline
V-Net&3&0&30.74&18.84&6.97&26.45\\
V-Net&60&0&81.46&69.18&1.31&5.09\\
\hline
MT~\cite{tarvainen2017mean}&3&57&31.09&18.77&28.14&59.22\\
\reviseagain{
DAN~\cite{DBLP:conf/miccai/ZhangYCFHC17}}&3&57&46.37&30.84&16.87&42.89\\
\reviseagain{Entropy Mini\cite{DBLP:conf/cvpr/VuJBCP19}}&3&57&50.71&35.13&16.14&42.45\\
UA-MT\cite{DBLP:conf/miccai/YuWLFH19}&3&57&34.46&21.24&25.73&57.40\\
\reviseagain{CCT~\cite{DBLP:conf/cvpr/OualiHT20}}&3&57&43.89&28.72&20.81&52.58\\
DTC~\cite{luo2021semi}&3&57&48.47&32.71&17.03&42.61\\
SASSNet~\cite{DBLP:conf/miccai/LiZH20}&3&57&51.96&36.03&16.08&45.36\\
\reviseagain{MC-Net~\cite{wu2022mutual}}&3&57&50.26&35.41&11.27&30.05\\
\reviseagain{URPC~\cite{luo2021urpc}}&3&57&55.00&39.23&\textbf{6.11}&\textbf{22.40}\\
\hline
Ours&3&57&\textbf{56.18}&\textbf{40.05}&12.47&34.85\\
\bottomrule[1.5pt]
\end{tabular}
\end{table*}


\section{Experimental Results}
\subsection{Datasets and Preprocessing}

\noindent \textbf{Pancreas CT dataset.} 
Pancreas CT dataset~\cite{DBLP:conf/miccai/RothLFSLTS15} is a public dataset containing 80 scans with a resolution of 512$\times$512 pixels and slice thickness between 1.5 and 2.5 mm. 
Each image has a corresponding pixel-wise label, which is annotated by an expert and verified by a radiologist. 

\noindent \textbf{Colon cancer segmentation dataset.}
Colon cancer dataset is a subset from Medical Segmentation Decathlon (MSD) datasets~\cite{simpson2019large}, consisting of 190 colon cancer CT volumes. Pixel-level label annotations are given on 126 CT volumes. Among these volumes, we randomly split 26 CT volumes as a test set and use the rest for training. 

\noindent \textbf{The Left Atrium (LA) MR dataset}. LA dataset contains 100 MR image scans with an isotropic resolution of $0.625\times 0.625 \times 0.625 mm^3$.
This dataset is fully annotated with pixel-level supervision, among which 80 scans are used for training and the remaining 20 are used for validation.

\noindent \textbf{Preprocessing.}
To fairly compare with other methods, we follow preprocessing in \cite{luo2021semi} by clipping CT images to a range of [$-$125, 275] HU values, resampling images to 1$\times$1$\times$1$mm$ resolution, center-cropping both raw images and annotations around foreground area with a margin of 25 voxels and finally normalizing raw images to zero mean and unit variance.
On the Pancreas dataset, we apply random crop as an augmentation on the fly, and the Colon dataset is augmented with random rotation, random flip, and random crop.
On both CT datasets, $96\times96\times96$ sub-volumes are randomly cropped from raw data and fed to the segmentation model for training.
On LA dataset, we apply center crop as well as normalizing the intensities to zero mean and unit variance for pre-processing.
During training, we adopt random crop to $112 \times 113 \times 80$ for on-the-fly augmentation.



\subsection{Implementation Details}
\label{sec:imp_details}
\noindent \textbf{Environment.}
All experiments in this work are implemented in Pytorch 1.6.0 and conducted under python 3.7.4 running on an NVIDIA TITAN RTX.

\noindent \textbf{Backbone.}
VNet \cite{DBLP:conf/3dim/MilletariNA16} is used as our backbone where the last convolutional layer is replaced by a 3D 1$\times$1$\times$1 convolutional layer.
On top of that, a projection module and an aleatoric uncertainty module are built for feature regularization and aleatoric uncertainty estimation, respectively.
Similar to \cite{DBLP:journals/corr/abs-2012-06985}, the projection head constitutes 3 convolutional layers, each followed by ReLU activations and batch normalization, except for the last layer, which is  followed by a unit-normalization layer. The channel size of each convolutional layer is set as 16. The aleatoric uncertainty module is comprised of three 1-layer branches predicting means, covariance factors, and covariance diagonals respectively.
\reviseagain{
Output sampling is implemented by calling the sample function of torch.distributions.LowRankMultivariateNormal~\footnote{\url{https://pytorch.org/docs/stable/distributions.html\#torch.distributions.lowrank_multivariate_normal.LowRankMultivariateNormal}}.
The teacher model parameters are updated by taking a moving average of the student model parameters. 
For BCL, the near-boundary pixels are obtained from the difference set of original foreground pixels and resulting foreground pixels after morphology dilation of 1-pixel radius.
}

\noindent \textbf{Training details.}
Our model is trained with an SGD optimizer with 0.9 momentum and 0.0001 weight decay for 6000 iterations.
A step decay learning rate schedule is applied where the initial learning rate is set to be 0.01 and dropped by 0.1 every 2500 iterations.
For each iteration, a training batch containing two labeled and two unlabeled sub-volumes is fed to the proposed model,
with each sub-volume randomly cropped with the size of 96$\times$96$\times$96 for CT volumes and 112$\times$112$\times$80 for MRI.
On the test set, 
predictions on sub-volumes with the same size using a sliding window strategy with a stride of 16$\times$16$\times$16 (for CT volumes) or 18$\times$18$\times$4
(for MRI on LA dataset)
are fused to obtain the final results.

\noindent \textbf{Evaluation metrics.}
We use Dice (DI), Jaccard (JA), the average surface distance (ASD), and the 95$\%$ Hausdorff Distance (95HD) to evaluate the effectiveness of our semi-supervised segmentation method. 
\revise{DI and JA mainly measure the amount of overlap between output segmentation maps and human annotations. The latter two metrics, ASD and 95HD, measure surface distance and are more sensitive to errors over the segmentation boundary.}
    

\begin{table*}[!ht]
\centering
\caption{A comparison with state-of-the-art on Colon tumor dataset with 5\% labeled data.}\label{tab:low_colon}
\begin{tabular}{c|c|c|c|c|c|c}
\toprule[1.5pt]
\multirow{2}{*}{Method} & \multicolumn{2}{c}{$\#$ scans used} & \multicolumn{4}{|c}{Metrics}\\
\cline{2-7}
&Labeled&Unlabeled&Dice[\%]$\uparrow$&Jaccard[\%]$\uparrow$&ASD[voxel]$\downarrow$&95HD[voxel]$\downarrow$\\
\hline
V-Net&5&0&34.07&23.09&10.12&26.52\\
V-Net&100&0&62.31&49.47&2.14&13.49\\
\hline
MT~\cite{tarvainen2017mean}&5&95&38.64&26.38&14.41&33.08\\
\reviseagain{
DAN~\cite{DBLP:conf/miccai/ZhangYCFHC17}}&5&95&34.02&24.14&14.21&32.42\\
\reviseagain{Entropy Mini~\cite{DBLP:conf/cvpr/VuJBCP19}}&5&95&38.62&26.78&18.02&39.01\\
UA-MT\cite{DBLP:conf/miccai/YuWLFH19}&5&95&40.61&28.01&15.31&34.92\\
\reviseagain{CCT~\cite{DBLP:conf/cvpr/OualiHT20}}&5&95&43.74&30.64&11.23&26.21\\
SASSNet~\cite{DBLP:conf/miccai/LiZH20}&5&95&41.64&30.07&11.93&28.96\\
DTC~\cite{luo2021semi}&5&95&43.29&29.84&10.62&26.22\\
\reviseagain{MC-Net~\cite{wu2022mutual}}&5&95&38.71&26.90&12.19&28.52\\
\reviseagain{
URPC~\cite{luo2021urpc}}&5&95&46.43&33.01&9.31&24.57\\
\hline
Ours&5&95&\textbf{49.00}&\textbf{35.15}&\textbf{9.04}&\textbf{22.32}\\
\bottomrule[1.5pt]
\end{tabular}
\end{table*}

\begin{table*}[!ht]
\centering
\caption{
A comparison with state-of-the-art on LA dataset with 20\% labeled data.}
\label{tab:cmp_la}
\reviseagain{
\begin{tabular}{c|c|c|c|c|c|c}
\toprule[1.5pt]
\multirow{2}{*}{Method} & \multicolumn{2}{c}{$\#$ scans used} & \multicolumn{4}{|c}{Metrics}\\
\cline{2-7}
&Labeled&Unlabeled&Dice[\%]$\uparrow$&Jaccard[\%]$\uparrow$&ASD[voxel]$\downarrow$&95HD[voxel]$\downarrow$\\
\hline
V-Net&16&0&86.03&76.06&3.51&14.26\\
V-Net&80&0&91.14&83.82&1.52&5.75\\
\hline
MT\cite{tarvainen2017mean}&16&64&88.42&79.45&2.73&13.07\\
DAN~\cite{DBLP:conf/miccai/ZhangYCFHC17}&16&64&87.52&78.29&2.42&9.01\\
Entropy Mini\cite{DBLP:conf/cvpr/VuJBCP19}&16&64&88.45&79.51&3.72&14.14\\
UA-MT~\cite{DBLP:conf/miccai/YuWLFH19}&16&64&88.88&80.21 &2.26&7.32\\
ICT~\cite{DBLP:conf/ijcai/VermaLKBL19}&16&64&89.02&80.34 &1.97&10.38\\
SASSNet~\cite{DBLP:conf/miccai/LiZH20}&16&64&89.27&80.82&3.13&8.83\\
DTC~\cite{luo2021semi}&16&64&89.42 &80.98&2.10&7.32\\
Chaitanya et al.~\cite{chaitanya2020contrastive}&16&64&89.94&81.82&2.66&7.23\\
SimCVD~\cite{you2022simcvd}&16&64&90.85&\textbf{83.80}&1.86&6.03\\
MC-Net~\cite{wu2022mutual}&16&64&91.07&83.67&\textbf{1.67}&5.84\\
\hline
Ours&16&64&\textbf{91.08}&83.67&1.80&\textbf{5.60}\\
\bottomrule[1.5pt]
\end{tabular}
}
\end{table*}

\subsection{Results on Pancreas Dataset}
\reviseagain{\textbf{Our settings}}. Since the predictions on unlabeled data may be  inaccurate in the early stage of training,
we follow common practices~\cite{DBLP:conf/miccai/YuWLFH19,luo2021semi} and use a Gaussian ramping up function 
$\lambda_g(t)=0.15*e^{-5(1-\frac{t}{t_{max}})^2}$
to control the strength of consistency regularization,
where $t$ denotes current time step and $t_{max}$
denotes the maximal training step, \ie, 6000 as introduced previously. The constant used to scale BCL, \ie, $\lambda_c$
is set to be 0.09 given $20\%$ labeled data and 0.01 given $5\%$ labeled data. In the second stage of training, the PCL weight $\lambda_r$ is always set to be 0.1.

\reviseagain{\textbf{Compared methods.}}
Table~\ref{tab:results_pancreas_cmp} shows the results on the pancreas dataset.
\reviseagain{
We compare with recent algorithms including mean teacher
(MT)~\cite{tarvainen2017mean}, 
Deep Adversarial Network (DAN)~\cite{DBLP:conf/miccai/ZhangYCFHC17}, 
Entropy Minimization
(Entropy Mini)~\cite{DBLP:conf/cvpr/VuJBCP19},
Uncertainty Aware Mean
Teacher 
(UAMT)~\cite{DBLP:conf/miccai/YuWLFH19},
cross-consistency training method 
(CCT)~\cite{DBLP:conf/cvpr/OualiHT20},
shapeaware adversarial network
(SASSNet)~\cite{DBLP:conf/miccai/LiZH20},
Dual-task Consistency
(DTC)~\cite{luo2021semi},
Uncertainty Rectified Pyramid Consistency (URPC)~\cite{luo2021urpc} and 
Mutual Consistency Network
(MC-Net)~\cite{wu2022mutual}.
}
Previous methods are mainly benchmarked on the first version of the Pancreas dataset with 12 labeled volumes and 50 unlabeled volumes, where, however, 2 duplicates of scan $\#$2
are found.
In case some of these three samples are in the training set and the rest are in the test set after a random split, we use version 2 where two duplicates are removed,
leaving us the same number of labeled data but 2 less, \ie, 48 unlabeled data.
Even under a more strict scenario, our proposed model achieves the best performance among existing works.


\reviseagain{\textbf{Results analysis.}}
The first row, \ie, a fully supervised baseline on the partial dataset, shows the lower bound of the semi-supervised segmentation methods. Whereas the second row, \ie, a fully supervised model on a fully labeled dataset, shows the upper bound performance.
We can observe that our method achieves 79.81\% on Dice, surpassing the current state-of-the-art by \reviseagain{0.54\%}.
Notably, our method is very close to the fully-supervised model that employs all volumes supervised by human annotations, showing the effectiveness of the proposed semi-supervised method.

\reviseagain{\textbf{A more challenging setting with 5\% labeled data.}}
To further validate our method under a more challenging scenario, we reduce the number of labeled data to only 5\% and use the rest 95\% as unlabeled. 
As shown in~Table~\ref{tab:low_pancreas},
in such a small-data regime, a performance drop of every semi-supervised learning method is observed compared to its counterpart in a big-data regime in Table~\ref{tab:results_pancreas_cmp} where 20\% labeled are available,
which confirms common sense.
It is observed that our method consistently outperforms other methods. 
Specifically, our method surpasses all the other semi-supervised methods and outperforms the current state-of-the-art by \reviseagain{1.18\% }on Dice, which demonstrates that the effectiveness of our method is more obvious in a more challenging setting. 

\reviseagain{\textbf{Comparison of computational cost.}
Due to a two-stage pipeline, our method takes around 2x hours to finish training compared with existing single-stage works. However, during inference, the proposed method does not introduce heavy computational overhead. Existing works use V-Net as the backbone for inference and their computational time costs are very similar. As mentioned in Section~\ref{sec:imp_details}, we only append one more layer after V-Net and the time cost is very close: 4.70 (ours) vs. 4.67 (V-Net), measured by seconds per sample. It means that in practical use, our method is as efficient as existing works.
}

\begin{table*}[!ht]
\centering
\caption{Ablation study on Pancreas dataset. BCL refers to \emph{boundary-aware contrastive learning} and PCL refers to \emph{prototype-aware contrastive learning}. Pseudo labeling refers to directly retraining the network with pseudo labels  \emph{without PCL}.}\label{tab:ablation_pancreas}
\begin{tabular}{c|c|c|c|c}
\toprule[1.5pt]
\multirow{2}{*}{Method}& \multicolumn{4}{c}{Metrics}\\
\cline{2-5}
&Dice$[\%]\uparrow$&Jaccard$[\%]\uparrow$&ASD[voxel]$\downarrow$&95HD[voxel]$\downarrow$\\
\hline
Supervised baseline&70.63&56.72&6.29&22.54\\
AUA
&76.13&62.19&2.25&9.35\\
AUA + BCL (First stage)
&77.15&63.34&2.04&7.00\\
AUA + BCL + Pseudo labeling
&79.08&65.91&1.91&6.69\\
AUA + BCL + Pseudo labeling + PCL (Full)
&\textbf{79.81}&\textbf{66.82}&\textbf{1.64}&\textbf{5.90}\\
\bottomrule[1.5pt]
\end{tabular}
\end{table*}

\begin{table*}[!ht]
\centering
\caption{Ablation study on Colon dataset. BCL refers to \emph{boundary-aware contrastive learning} and PCL refers to \emph{prototype-aware contrastive learning}. Pseudo labeling refers to directly retraining the network with pseudo labels  \emph{without PCL}.}
\label{tab:ablation_colon}
\begin{tabular}{c|c|c|c|c}
\toprule[1.5pt]
\multirow{2}{*}{Method}& \multicolumn{4}{c}{Metrics}\\
\cline{2-5}
&Dice$[\%]\uparrow$&Jaccard$[\%]\uparrow$&ASD[voxel]$\downarrow$&95HD[voxel]$\downarrow$\\
\hline
Supervised baseline&34.07&23.09&10.12&26.52\\
AUA
&42.74&30.20&15.00&35.43\\
AUA + BCL (First stage)
&43.70&30.92&14.74&33.34\\
AUA + BCL + Pseudo labeling
&46.75&33.62&12.39&28.49\\
AUA + BCL + Pseudo labeling + PCL (Full)
&\textbf{49.00}&\textbf{35.15}&\textbf{9.04}&\textbf{22.32}\\
\bottomrule[1.5pt]
\end{tabular}
\end{table*}

\begin{figure*}[!t]
\centering
\includegraphics[width=1\textwidth]{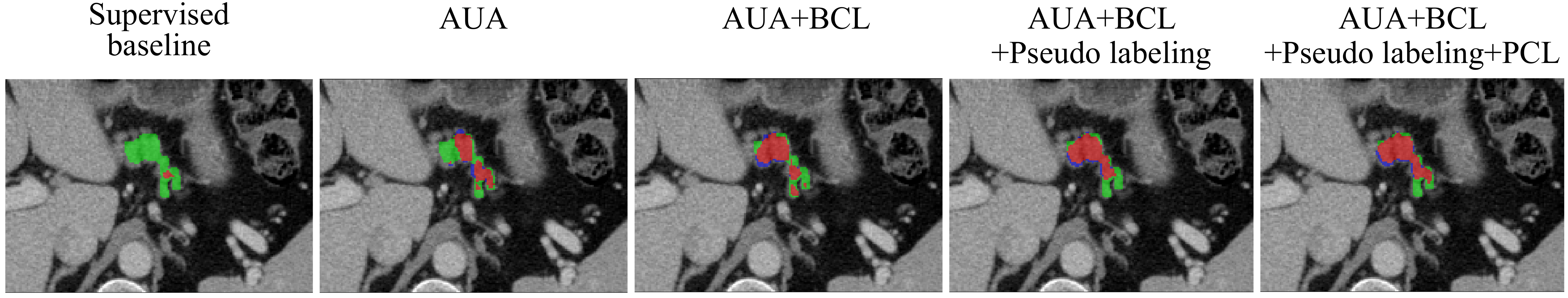}

\caption{
\reviseagain{A visualized ablation study.
Regions highlighted in red are true positive areas, \ie, pixels correctly predicted.
Green and blue regions represent false negatives, \ie, foreground pixels incorrectly predicted as background, and false positives, \ie, 
background pixels incorrectly predicted as foreground.
}
}
\label{fig:vis_abl}
\end{figure*}

\subsection{Results on Colon Dataset}

Table~\ref{tab:low_colon} shows the results on the colon dataset. To get a result, we set
$\lambda_g(t)$ to be $0.15*e^{-5(1-\frac{t}{t_{max}})^2}$,
and the scaling weight of BCL, \ie, $\lambda_c$ is set to be 0.03.
In the second stage of training, the weight PCL is set to be 0.1.
We compare our method
with several state-of-the-art methods 
using 
$5\%$ data as labeled and the rest as unlabeled.
Again, we tune hyper-parameters for previous methods
so that these methods can reach the best performance on this dataset.
In Table~\ref{tab:low_colon}, 
by comparing the second row with Table~\ref{tab:low_pancreas},
we notice under a fully supervised setting using a full dataset, 
the performance on the Colon dataset is lower than Pancreas dataset,
indicating Colon dataset is more challenging.
By comparing semi-supervised segmentation methods with a fully supervised setting using the partial dataset, \ie, the result in the first row of Table~\ref{tab:low_colon}, 
we observe stronger performance, showing that leveraging unlabeled data can improve the segmentation performance,
which confirms common sense.
Our method achieves superior performance compared with all previous works by a large margin \reviseagain{($3.43\%$)},
which indicates that our method can make better use of unlabeled data.

\reviseagain{
\subsection{Results on LA Dataset}
To demonstrate that our method is generalizable to different medical image modalities, we also conduct comparative experiments on the LA dataset, as shown in Table~\ref{tab:cmp_la}.
To get a result, we set
$\lambda_g(t)$ to be $0.15*e^{-5(1-\frac{t}{t_{max}})^2}$,
and the scaling weight of BCL, \ie, $\lambda_c$ is set to be 0.09.
In the second stage of training, the weight PCL is set to 0.1.
We compare with state-of-the-arts benchmarked in~\cite{you2022simcvd}
using $20\%$ data as labeled and the rest as unlabeled.
As a sanity check, all semi-supervised methods can outperform a fully-supervised baseline on the partial dataset (\ie, 16 labeled MR images),
demonstrating a meaningful utilization of unlabeled data.
In Table~\ref{tab:cmp_la}, 
compared with previous works,
the proposed method achieves the best results, closing the performance gap with a fully-supervised upper bound (\ie, trained with a full dataset).
}

\subsection{Ablation Studies}
Here we ablate each component of our proposed framework on the Pancreas dataset with $20\%$ as labeled (Table~\ref{tab:ablation_pancreas}) and on the Colon dataset with $5\%$ as labeled (Table~\ref{tab:ablation_colon}).
We gradually add our proposed component and showcase their performances in terms of four metrics.

\reviseagain{
Firstly, we validate the effectiveness of the adaptive supervision fitting scheme: AUA.
On both datasets, as shown in the second row of Table~\ref{tab:ablation_pancreas} and~\ref{tab:ablation_colon},
applying AUA achieves superior performance over the fully-supervised model, \ie, V-Net.
This performance gain mainly comes from its ability to identify and adaptively learn from trustworthy supervision. 
Specifically, AUA automatically estimates aleatoric uncertainty of input data and down-weights supervision from low-quality images, 
so that the student model learns from more accurate supervision of the teacher model.
Secondly, we demonstrate the effectiveness our stage-aware contrastive learning.
BCL can boost performance on top of AUA further by a margin of 1.02\% and 0.96\% on Pancreas and Colon datasets, respectively,
and PCL improves its baseline (as shown in the fourth row) by 0.73\% and 2.25\%, respectively.
Both techniques are designed to pull features of pixels belonging to the same class closer and push features belonging to opposite classes further, which shapes a more compact (inside each class) and better separated (across different classes) feature space,
leading to a more robust and effective semi-supervised method.
}

\reviseagain{
To get a qualitative sense of the effectiveness of each component of our pipeline, in Figure~\ref{fig:vis_abl}, we plot segmentation results by gradually adding the proposed technique.
We illustrate a failure case of the baseline model where the foreground pixels are mislabeled as background.
Adding our technique one by one is able to recall more foreground pixels and output segmentation maps with gradually larger overlap with the ground truth.

Additionally, we ablate the effect of 
$\lambda_c$,
the weight balancing AUA loss and BCL loss in the first stage of training.
Its values are chosen from 0.03, 0.05, 0.07, and 0.09.
Table~\ref{tab:ablation_lambda_c} demonstrates that the final results are robust to various choices of $\lambda_c$.
}

\begin{figure*}[!t]
\centering
\includegraphics[width=1\textwidth]{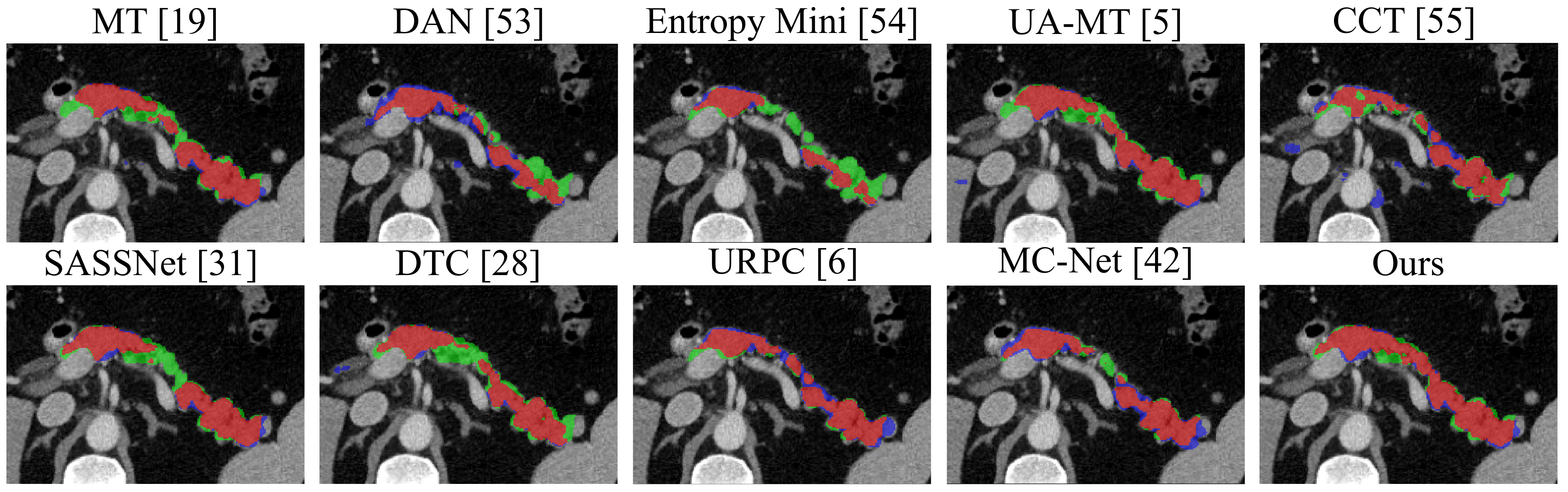}
\caption{\reviseagain{A comparison between visualized segmentation maps obtained by the state-of-the-art
and our method.
Regions highlighted in red are true positive areas, \ie, pixels correctly predicted.
Regions highlighted in green and blue are false negatives, \ie, foreground pixels incorrectly predicted as background, and false positives, \ie, 
background pixels incorrectly predicted as foreground.}
}
\label{fig:vis_cmp}
\end{figure*}

\begin{table}[!th]
\centering
\caption{
Ablation study on the effect of hyperparameter $\lambda_c$ on the Pancreas dataset.}
\reviseagain{
\begin{tabular}{c|c|c|c|c}
\toprule[1.5pt]
\multirow{2}{*}{Method}& \multicolumn{4}{c}{Metrics}\\
\cline{2-5}
&Dice$[\%]\uparrow$&Jaccard$[\%]\uparrow$&ASD[voxel]$\downarrow$&95HD[voxel]$\downarrow$\\
\hline
0.03
&79.26&66.03&1.84&6.45\\
0.05
&79.61&66.49&2.03&7.53\\
0.07
&79.45&66.31&2.04&6.44\\
0.09
&\textbf{79.81}&\textbf{66.82}&\textbf{1.64}&\textbf{5.90}\\
\bottomrule[1.5pt]
\end{tabular}}
\label{tab:ablation_lambda_c}
\end{table}

\reviseagain{
Finally, we ablate the robustness of the proposed method to the amount of unlabeled data.
in~\ref{tab:ablation_number_unlable}, we demonstrate the performance of our method by increasing the unlabeled data number from a small split (\ie, 1/4) to full. We can observe a growing trend in performance, and it is safe to conjecture that with more in-domain unlabeled data, our method can obtain extra performance gain.
}

\begin{table}[!th]
\centering
\caption{Ablation study on an increasing number of unlabeled data.}
\label{tab:ablation_number_unlable}
\scalebox{0.95}{
\reviseagain{
\begin{tabular}{c|c|c|c|c}
\toprule[1.5pt]
\multirow{2}{*}{\#Unlabeled}& \multicolumn{4}{c}{Metrics}\\
\cline{2-5}
&Dice[\%]$\uparrow$&Jaccard[\%]$\uparrow$&ASD[voxel]$\downarrow$&95HD[voxel]$\downarrow$\\
\hline
12 (1/4)
&76.77&62.95&1.76&8.47\\
24 (1/2)
&78.63&65.37&1.81&6.79\\
48 (Full)
&\textbf{79.81}&\textbf{66.82}&\textbf{1.64}&\textbf{5.90}\\
\bottomrule[1.5pt]
\end{tabular}
}
}
\end{table}

\subsection{A Qualitative Comparison with the State-of-the-arts}
We visualize the segmentation predictions obtained from other state-of-the-art methods and ours in Figure \ref{fig:vis_cmp}. 
We highlight true positive, false negative, and false positive pixels in red, green, and blue, respectively.
We can observe that for the other state-of-the-art works, 
they either achieve a lower recall, such as MT~\cite{tarvainen2017mean}, 
DAN~\cite{DBLP:conf/miccai/ZhangYCFHC17}, 
Entropy Mini~\cite{DBLP:conf/cvpr/VuJBCP19},
UA-MT~\cite{DBLP:conf/miccai/YuWLFH19},
SASSNet~\cite{DBLP:conf/miccai/LiZH20},
DTC~\cite{luo2021semi} and 
MC-Net~\cite{wu2022mutual}, or suffer from more false positives, such as DAN~\cite{DBLP:conf/miccai/ZhangYCFHC17} and URPC~\cite{luo2021urpc}.
However, the prediction of our method has a greater overlap with the ground truth.
\section{Discussion}

\reviseagain{
In this paper, to alleviate heavy reliance on pixel-wise labels, which requires considerable human efforts, we propose a novel semi-supervised learning method by taking advantage of aleatoric uncertainty estimation and exploring feature representation learning.
Firstly, on top of the mean teacher framework, we present AUA that estimates each image's aleatoric uncertainty and automatically down-weights supervision on ambiguous images so that the trained model is able to generate more reliable pseudo labels.
However, image ambiguity is an under-explored aspect in semi-supervised learning.
Secondly, we explore representation learning and propose a state-adaptive contrastive learning method.
In the first stage, a boundary-aware contrastive loss is designed to regularize labeled image features and in the second stage,
we use a prototype-aware contrastive loss to regularize both labeled and unlabeled features.
Superior performance across Pancreas-CT, Colon cancer, and LA datasets validates the superior performance of our method as well well its generality to different data modalities.

This study has comprehensive applicability.
Firstly, the proposed method can be used to automate downstream diagnosis. It is found in recent research~\cite {wu2022seatrans,zhou2019collaborative} that combining segmentation results benefits disease diagnosis classification tasks. The proposed method allows for training a segmentation model that achieves satisfactory results without relying on large-scale human labels and thus can be applied to downstream disease diagnosis tasks. Secondly, in clinical practice, our method can serve as another expert for doctors’ reference. For example, a doctor may take into consideration colon cancer segmentation results of the proposed method and make a better diagnosis of cancer staging. 

The main limitation of this study is lacking an automatic mechanism to differentiate incorrect pseudo labels from correct ones in the second stage. In this work, we put more effort into generating more accurate pseudo labels prior to their use. But given pseudo labels, how to make better use is also a non-trivial question. Online confidence thresholding~\cite{cascante2021curriculum,guo2022class} can be a potential solution to identify noisy pseudo labels out of clean ones. In addition, developing a more noisy label-tolerant loss function on our design could also get an extra performance gain.
}

\section{Conclusion}

\reviseagain{
This paper presents a simple yet effective two-stage framework for semi-supervised medical image segmentation, with the key idea of exploring the feature representation from labeled and unlabeled images. We propose a stage-adaptive contrastive learning method, including a boundary-aware contrastive loss and a prototype-aware contrastive loss. In the first stage, BCL loss regularizes features by pulling features sharing the same labels closer and pushing features with opposite labels further, arriving at a more compact and well-separated feature space. This loss function, together with AUA, which adaptively encourage consistency by considering the ambiguity of medical images, enhances pseudo label quality after the first stage of training. Improved pseudo labels not only provide higher quality supervision for the segmentation head but also generate more accurate prototypes, which allows PCL to regularize a well-separated feature space further. Specifically, the feature of each pixel is pulled closer to its class centroid and pushed away from its opposite class centroid, which translates to a more accurate segmentation model. Our method achieves the best results on three public medical image segmentation benchmarks,}
and the ablation study validates the effectiveness of our proposed method. Our future works include extending this work to different types of medical data, such as X-ray images, fundus images, and surgical videos.



\bibliographystyle{ieeetr}

\small{\bibliography{refs}}

\end{document}